\documentclass{article}

\usepackage{arxiv}

\usepackage[utf8]{inputenc} % allow utf-8 input
\usepackage[T1]{fontenc}    % use 8-bit T1 fonts
\usepackage{hyperref}       % hyperlinks
\usepackage{url}            % simple URL typesetting
\usepackage{booktabs}       % professional-quality tables
\usepackage{amsfonts}       % blackboard math symbols
\usepackage{nicefrac}       % compact symbols for 1/2, etc.
\usepackage{microtype}      % microtypography
\usepackage{lipsum}
\usepackage{graphicx}
\usepackage{times}
\usepackage{multirow}
\usepackage{amsmath}
\usepackage{tikz}
\graphicspath{ {./images/} }

\title{Multimodal Model for Computational Pathology:\\
Representation Learning and Image Compression}

\author{{\hspace{1mm}Peihang Wu}\thanks{Peihang Wu and Zehong Chen contributed equally to this work.} \\
	\textsuperscript{1}Shenzhen University of Advanced Technology\\
	\And
	\hspace{1mm}Zehong Chen\textsuperscript{*} \\
	\textsuperscript{1}Shenzhen University of Advanced Technology\\
    \And
	\hspace{1mm}Lijian Xu\thanks{Corresponding author.} \\
	Shenzhen University of Advanced Technology \\
	\texttt{xulijian@suat-sz.edu.cn} \\
}

\begin{document}
\maketitle

\begin{abstract}

Whole slide imaging (WSI) has revolutionized digital pathology by enabling large-scale computational analysis of histopathological images at gigapixel resolution. Recent advances in foundation models have significantly accelerated progress in computational pathology. These models enable joint reasoning across heterogeneous medical modalities, including pathology images, clinical reports, and structured biomedical data.
Despite rapid progress, several challenges remain in developing scalable and reliable artificial intelligence systems for clinical pathology. First, the extremely high spatial resolution of WSIs introduces substantial computational challenges for visual representation learning. Second, limited availability of expert annotations restricts the development of supervised learning approaches. Third, integrating multimodal information while maintaining biological consistency and interpretability remains an open research problem. Fourth, the opaque nature of modeling ultra-long visual sequences in gigapixel WSIs hinders the clinical transparency and reliability required for decision-making.
This review provides a comprehensive overview of recent advances in multimodal computational pathology. We systematically analyze four key research directions: 
(1) self-supervised representation learning and structure-aware token compression for whole-slide images; 
(2) multimodal data generation and augmentation; 
(3) parameter-efficient adaptation and reasoning-enhanced few-shot learning; 
and (4) multi-agent collaborative reasoning for trustworthy diagnosis.
We specifically analyze how token compression enables cross-scale modeling and how multi-agent mechanisms simulate a pathologist's "Chain of Thought" across magnifications to achieve uncertainty-aware evidence fusion.
Finally, we discuss open challenges and argue that future progress depends on unified multimodal frameworks that integrate high-resolution visual data with clinical and biomedical knowledge to support interpretable and safe AI-assisted diagnosis.


\end{abstract}

\section{Introduction}

\begin{figure}[t]
    \centering
    \includegraphics[width=0.85\textwidth]{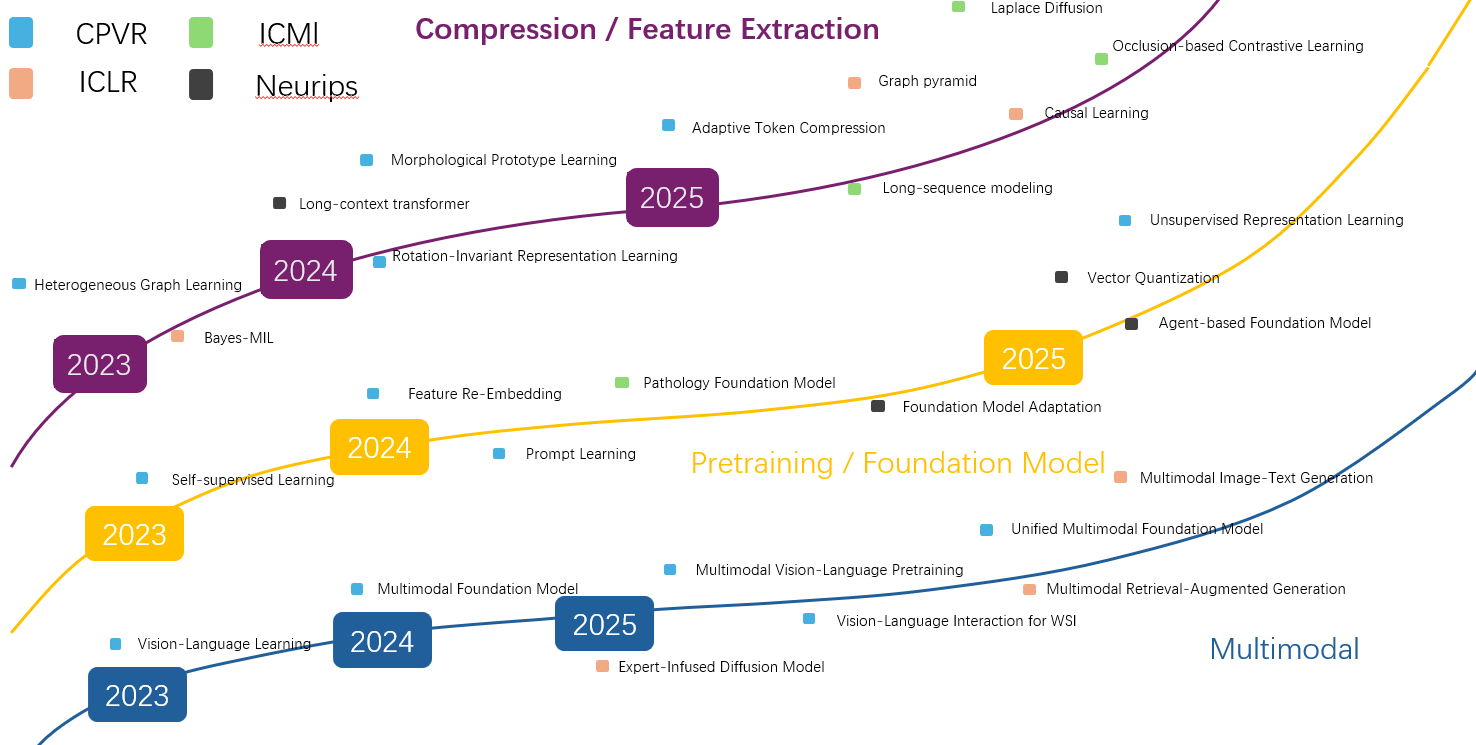}
    \caption{Timeline of milestone AI models in computational pathology (2021–2025). The chart illustrates the chronological progression of key literature reviewed in this survey. Models are positioned according to their publication or preprint release dates, capturing the rapid evolution from foundation models to recent advanced multimodal frameworks. }
    \label{fig:timeline}
\end{figure}

Whole slide imaging (WSI) has become a cornerstone of digital pathology, enabling computational analysis of histopathological images at gigapixel resolution. The growing demand for intelligent clinical decision support has spurred interest in multimodal learning approaches that jointly analyze pathology images and textual clinical information. As illustrated in Figure~\ref{fig:timeline}, a surge of influential works has emerged in computational pathology over recent years.

Deep learning has achieved remarkable success in visual representation learning and has been widely adopted in medical image analysis. 
In computer vision, large-scale self-supervised learning paradigms such as contrastive learning and masked modeling have significantly improved representation learning without requiring extensive manual annotations \cite{Chen2020SimCLR,He2022MAE}. 
These approaches have gradually been introduced into computational pathology, enabling models to learn generalizable representations from large collections of WSIs \cite{Lu2021DataEfficientCPath,Xu2024WholeSlideFM,Chen2024CPathFoundation}. 
Meanwhile, the emergence of pathology foundation models has further advanced this direction by leveraging large scale pretraining to improve cross task generalization and robustness \cite{Huang2023TwitterVLM,Lu2024CPathVLM,Tang2024Feature}.

In parallel, the rapid development of multimodal large language models has substantially enhanced the capability of AI systems to perform cross modal perception and reasoning. 
By integrating powerful language models with high capacity visual encoders, models such as GPT-5, DeepSeek-R1, Qwen3-VL, LLaVA, and BLIP-2 demonstrate strong performance in multimodal reasoning, document understanding, and visual question answering tasks \cite{Guo2025DeepSeekR1,Bai2025Qwen3VL,Singh2025GPT5SystemCard,Liu2023VisualInstructionTuning,Li2023BLIP2}. 
Inspired by these advances, recent studies have begun to explore multimodal foundation models tailored for pathology image analysis \cite{Lu2024PathologyCopilot,Chen2025InstructionPathology,Sun2025CPathOmni}. 
By jointly modeling histopathological images and diagnostic reports, these approaches aim to achieve richer semantic understanding and improved clinical interpretability.

\begin{figure}[t]
    \centering
    \includegraphics[width=0.85\textwidth]{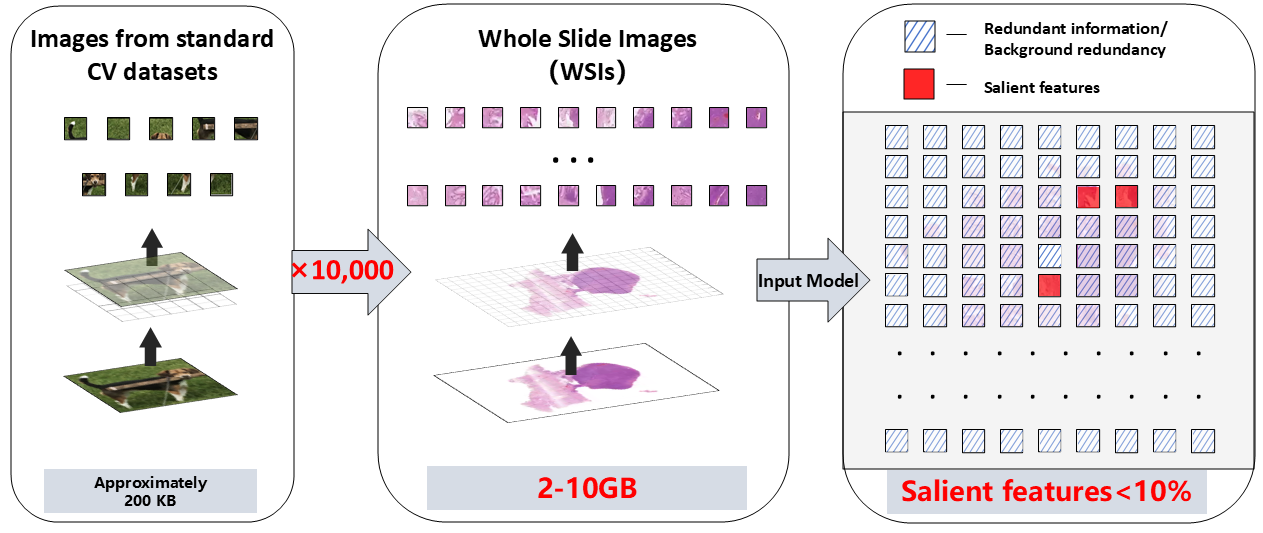}
    \caption{Schematic illustration of the data scale difference between pathological whole slide images and natural images. WSIs typically contain billions of pixels, imposing extremely high computational and storage demands. Moreover, due to tissue heterogeneity and uneven lesion distribution, diagnostically critical regions are often sparsely scattered within the vast background.}
\label{fig:diff_nature_path}
\end{figure}

WSIs typically contain billions of pixels, and their ultra high resolution imposes substantial computational and memory burdens during processing. In real clinical scenarios, due to tissue heterogeneity and uneven lesion distribution, diagnostically critical regions are often sparsely distributed (see Figure~\ref{fig:diff_nature_path}). Although patch-based processing alleviates computational cost to some extent, it disrupts global spatial structure and limits the model’s ability to capture complex pathological patterns holistically. Therefore, how to effectively model both global context and critical diagnostic evidence under limited computational budgets remains a central challenge in computational pathology.
Despite recent advances, pathology images exhibit characteristics that fundamentally distinguish them from natural images used in conventional computer vision. They lack a canonical orientation, as cellular and tissue structures remain semantically consistent under arbitrary rotations. Their color distribution is relatively concentrated and largely governed by hematoxylin and eosin staining patterns. In addition, their interpretation is inherently scale dependent, since the same tissue region may convey different biological meanings at different magnifications.

Another important challenge arises from the extremely high spatial resolution of WSIs. 
A single slide may contain billions of pixels and millions of cells, resulting in extremely long visual token sequences when processed by modern vision transformers. 
Traditional patch based approaches alleviate computational burdens by dividing WSIs into small image patches. 
However, this strategy disrupts global spatial continuity and restricts the ability of models to capture long-range contextual relationships across tissue structures. 
Recent studies have attempted to address these issues through hierarchical representation learning and cross scale modeling strategies \cite{Guo2025ContextMatters,li2024rethinking}. 
Meanwhile, several works have explored token compression and efficient transformer architectures to improve computational efficiency when processing long visual sequences \cite{Fayyaz2022TokenSampling,Zeng2024TCFormer,Wang2024TokenComplement,Ye2025VocoLLaMA,Bolya2022TokenMerging}. 
Nevertheless, these approaches often assume that redundant regions can be safely discarded, which may not hold for pathology images where diagnostically critical regions are sparse yet essential.

Beyond computational challenges, the scarcity of high quality annotated medical data further limits the development of robust pathology AI systems. 
Unlike natural image datasets, pathology annotation requires expert pathologists and is therefore extremely costly and time consuming. 
In addition, many disease subtypes follow long tail distributions, which further increases the difficulty of model training. 
To alleviate these limitations, recent studies have explored multimodal data synthesis and generative modeling techniques for data augmentation. 
Diffusion-based generative models have demonstrated promising capabilities in synthesizing realistic medical images \cite{Ding2023SyntheticPathologyDataset,Zhang2022PseudoHealthy,Li2025TopoFM}. 
Collaborative generation frameworks and multi-agent systems have also been proposed to generate large scale pathology image and text pairs for multimodal training \cite{Sun2025PathGen,Sun2025CPathAgent}. 
Meanwhile, enabling rapid model adaptation under limited supervision has become another important research direction. 
Recent studies have investigated multimodal transfer learning, prompt based learning, and adaptive fine tuning strategies to improve model adaptability in low data scenarios \cite{guo2025focus,Han2025MSCPT,huang2024free,yin2024prompting,Quan2024LabelSelection}. 
Other works have explored self-supervised learningapproaches to enhance cross domain generalization \cite{Ouyang2022SelfSupervisedSeg,Huang2023VectorQuantizationSeg,Kim2023MetaKronecker,Tong2025SelfDisentanglement}.

Finally, reliability and interpretability remain critical requirements for deploying AI systems in real clinical environments. 
Modern deep neural networks often operate as black box models, making it difficult for clinicians to understand the reasoning process behind diagnostic predictions. 
Recent studies have therefore explored uncertainty estimation and robust multimodal representation learning to improve the reliability of AI systems \cite{You2024CalibratingMultimodal,Huang2024AnomalyDetection}. 
Other works have investigated agent based reasoning frameworks that simulate the diagnostic workflow of human pathologists, enabling interpretable multi step analysis of pathology images \cite{Sun2025CPathAgent,Wu2025MoEWSI,Wang2026WSISum}. 
Beyond visual representation learning, downstream clinical tasks such as \textbf{survival prediction} \cite{zhou2025robust,liu2024interpretable} and \textbf{gene expression inference} \cite{nishimura2025learning,huang2025scalable} have also emerged as important benchmarks for evaluating the practical utility of multimodal pathology models.

In summary, the existing literature highlights several important directions for future research.
First, efficient self-supervised representation learning methods are required for ultra high resolution pathology images.
Second, biologically consistent multimodal data synthesis frameworks are needed to mitigate the scarcity of annotated clinical data.
Third, advanced model adaptation techniques should be developed to address long tail clinical tasks under limited supervision, leveraging reasoning and structured knowledge to maximize data efficiency.
Finally, reliable multimodal reasoning mechanisms are essential for achieving trustworthy and interpretable clinical decision support.
Advancing these directions will play a crucial role in building scalable and clinically applicable AI systems for digital pathology.

\section{Representation Learning for Computational Pathology}

Computational pathology aims to learn robust representations from gigapixel WSIs to support diagnosis, subtyping, prognosis prediction, and multimodal reasoning. The extreme resolution of WSIs, sparse supervision, and multimodal clinical context make representation learning particularly challenging. This section systematically reviews the evolution of supervision paradigms and the emergence of large-scale foundation models.

\subsection{Supervision Paradigms: From Multiple Instance Learning to Self-Supervision}

\textbf{Multiple instance learning (MIL)} remains the dominant paradigm for WSI analysis, where a slide is treated as a bag of image patches and patch-level representations are aggregated to produce slide-level predictions. TransMIL models dependencies across instances using transformer-based correlated attention, significantly improving contextual aggregation beyond independent pooling strategies \cite{Shao2021TransMIL}. However, a persistent limitation of conventional MIL frameworks lies in their reliance on frozen offline feature extractors. To bridge this gap, Feature Re-Embedding with a Re-embedded Regional Transformer (R2T) refines instance features online within the MIL pipeline, bringing conventional encoders closer to foundation-model-level representations \cite{Tang2024Feature}. Furthermore, ABMILX revisits optimization challenges in sparse-attention MIL under end-to-end learning, introducing global correlation-based refinement with multi-head attention to surpass two-stage pipelines while maintaining computational efficiency \cite{Tang2025Revisiting}.

Beyond single-modality visual modeling, \textbf{Multimodal MIL frameworks} extend slide modeling to heterogeneous clinical data. For instance, a gated mixture-of-experts architecture has been proposed to integrate WSI features with flow cytometry data, explicitly modeling superclass–subclass hierarchies to improve interpretability \cite{Hashimoto2024Multimodal}. Similarly, I2MoE introduces an interpretable interaction-aware mixture-of-experts framework that models complex multimodal interactions, providing both local and global transparency \cite{Xin2025IMoE}.

To address the high cost and time-intensive nature of dense expert annotations, \textbf{Weakly and Semi-Supervised Dense Representation Learning} have emerged as critical research directions. HAMIL performs weakly supervised segmentation using high-resolution activation maps and interleaved learning to generate reliable pseudo labels \cite{Zhong2023HAMIL}. Complementarily, CDMA+ introduces a semi-supervised segmentation framework incorporating cross-decoder knowledge distillation and Seg-CAM consistency to effectively leverage unlabeled data \cite{Zhong2024SemiSupervised}. These approaches focus on fine-grained spatial representation learning, effectively complementing global slide-level modeling paradigms.

\begin{figure}[htbp]
    \centering
    \includegraphics[width=0.85\textwidth]{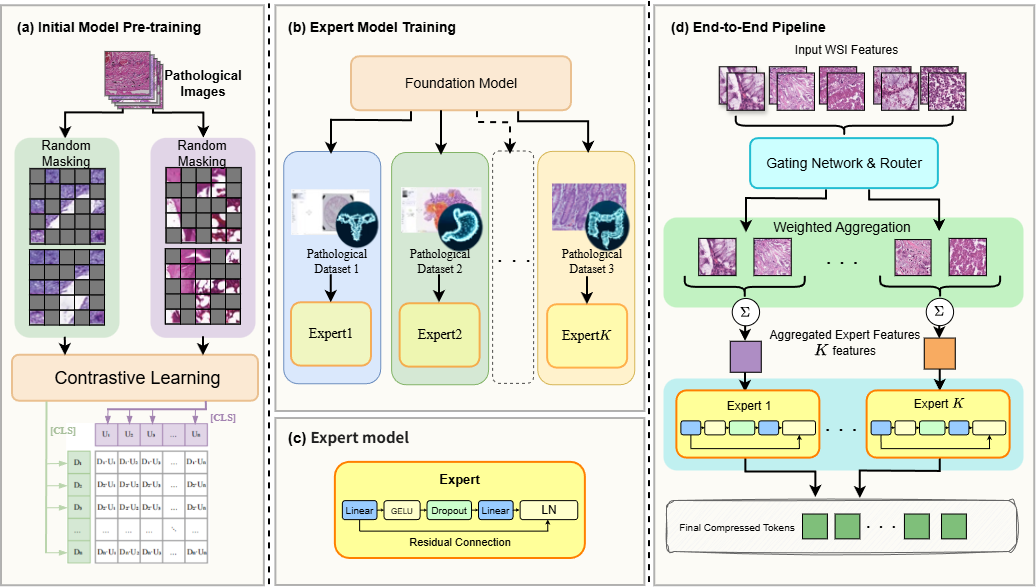}
    \caption{Schematic of a multi-task self-supervised learning framework for pathological images. The framework jointly optimizes three pretext tasks: (a) masked image modeling to reconstruct masked patches, enabling the model to learn tissue structure and context; (b) instance-level contrastive learning that pulls augmented views of the same image together while pushing different views apart, capturing discriminative morphological features; and (c) cross-resolution consistency learning to enforce feature invariance across magnifications. Combined pretraining yields a general-purpose pathology foundation model with strong diagnostic discriminability.}
\label{fig:selfsupervise}
\end{figure}

\textbf{Self-Supervised Representation Learning} has further established itself as a cornerstone for scalable representation learning under limited annotations. Figure~\ref{fig:selfsupervise} illustrates a conceptual overview of self-supervised learning paradigms in computational pathology.
This section tackles limited annotations and large token scale via joint self-supervised learning and token compression. Multi-task objectives, including masked modeling, contrastive learning, and cross-resolution consistency, learn robust representations. The model is then specialized into domain-specific experts and integrated with sparse gating for efficient aggregation. Finally, structure-aware importance estimation enables dynamic token compression, preserving critical regions while reducing redundancy.

Self-Supervised Representation Distribution Learning (SSRDL) introduces online representation sampling to learn distributions of patch embeddings rather than single invariant features, improving MIL robustness \cite{10643565}. 
Various self-supervised pretext tasks have been explored, including colorization and cross-channel reconstruction for survival prediction \cite{9980424}, as well as source-free open-set domain adaptation via Distill-SODA \cite{10403873}. 
Additionally, self-supervised disentanglement networks enable arbitrary stain transfer by separating tissue content from stain representations, enhancing cross-center generalization \cite{10192441}. 
Recent generative pretext tasks, such as GenSelfDiff-HIS, leverage diffusion-based image-to-image modeling for histopathological segmentation \cite{10663482}. 
For specialized modalities, S4R introduces separated self-supervised spectral regression for hyperspectral pathology imaging, enabling effective classification by reconstructing spectral bands \cite{11026788}. 
Furthermore, \textbf{Multi-Resolution Pathology-Language pretraining} \cite{albastaki2025multi,juyal2024pluto} aligns visual and textual representations across multiple resolutions of WSIs, allowing models to capture both fine-grained and global tissue features simultaneously, which enhances representation robustness and downstream generalization.

\subsection{Pathology Foundation Models and Multimodal Large Language Models}

The paradigm shift toward large-scale pretraining has catalyzed the emergence of pathology foundation models capable of universal task adaptation. UNI performs self-supervised pretraining on over 100 million pathology images from 100,000 WSIs, demonstrating strong transferability across 34 downstream tasks and introducing resolution-agnostic modeling \cite{Chen2024CPathFoundation}. Pushing the boundaries of scale, Prov-GigaPath utilizes 1.3 billion tiles to introduce architectures capable of modeling ultra-long slide contexts, achieving state-of-the-art performance in cancer subtyping and pathomics \cite{Xu2024WholeSlideFM}. These works represent a transition from patch-centric analysis to holistic slide-centric representation learning.

Histopathological diagnosis naturally integrates visual evidence with textual clinical knowledge, making visual-language pretraining a critical extension. PLIP performs visual–language contrastive learning on large image–text collections to enable zero-shot classification and retrieval \cite{Huang2023TwitterVLM}, while CONCH scales caption-based contrastive learning to over one million pairs, achieving state-of-the-art results in segmentation and captioning \cite{Lu2024CPathVLM}. Instruction-tuned assistants have further enhanced these capabilities; PathChat functions as a multimodal pathology copilot trained on visual-language instructions for improved clinical reasoning \cite{Lu2024PathologyCopilot}. CPath-Omni unifies patch-level and slide-level tasks within a single 15B-parameter model, supporting tasks from VQA to referring expression recognition \cite{Sun2025CPathOmni}.

However, modeling gigapixel slides in multimodal LLMs introduces significant computational challenges. LoC-Path addresses this by reducing redundancy through sparse token merging and cross-attention routing \cite{Hu2025LoCPath}. The broader biomedical context is supported by models like LLaVA-Med, which utilizes curriculum-based training \cite{Li2023LLaVAMed}, and MedDr, which employs diagnosis-guided bootstrapping for dataset construction \cite{He2024MedDr}. System-level modeling principles, as discussed in GPT-4o \cite{OpenAI2024GPTo}, continue to influence the field. Finally, text-guided representation learning strategies inject structured biomedical knowledge into vision encoders to improve data-efficient classification, ensuring that models benefit from both visual patterns and linguistic expertise \cite{Zhang2023TextGuided}.

\section{Advances in Whole-Slide Image Compression for Computational Pathology}

The rapid adoption of digital pathology has led to the routine generation of gigapixel whole-slide images, each containing tens of thousands to hundreds of thousands of high-resolution patches. While WSIs eliminate the need for physical slide storage, they introduce severe challenges in data storage, transmission, and large-scale model training. Consequently, compression has emerged as a central research theme spanning pixel-level coding, representation-level token reduction, semantic summarization, and multimodal knowledge-guided filtering.

\subsection{Pixel-Level and Structural Compression}

Traditional and learned \textbf{Pixel-Level Compression} methods aim to reduce the raw data footprint while preserving diagnostic fidelity. WISE introduces a hierarchical lossless encoding framework tailored for gigapixel pathology images, achieving compression ratios up to 136$\times$ through optimized coding \cite{mao2025wise}. Moving beyond static ratios, AdaSlide dynamically determines region-specific compression levels using reinforcement learning and integrates a foundational image enhancer to maintain downstream performance \cite{lee2025adaptive}. Furthermore, recent learned pathology image compression frameworks have demonstrated superior rate–distortion performance by specifically preserving fine-grained tissue morphology that is often lost in general-purpose codecs \cite{lee2025learned}.

Building on raw pixel efficiency, \textbf{Multi-Resolution and Segmentation-Aware Compression} leverages the inherent hierarchical structure of WSIs. WSI-SAM extends the Segment Anything Model to multi-resolution pathology images using a dual mask decoder to process high- and low-resolution tokens simultaneously \cite{liu2024wsi}. This structural awareness is further refined in multi-resolution pathology-language models, which align visual and textual representations across scales to improve both generalization and representation compactness, ensuring that the most informative scales are prioritized for storage and analysis \cite{albastaki2025multi}.

\subsection{Representation-Level and Implicit Compression}

At the feature level, \textbf{Representation-Level Compression} focuses on reducing the sequence length of tokens processed by vision transformers. DTC-WSI integrates saliency estimation and progressive token merging to reduce token counts without sacrificing classification accuracy \cite{rahman2026dtc}. FOCUS further enhances this by introducing a knowledge-guided framework that utilizes language priors from pathology foundation models to identify and retain only diagnostically relevant patches \cite{guo2025focus}. In an extreme case of summarization, WSISum formulates WSI analysis as a semantic reconstruction problem, proving that less than 2.5\% of patches can retain over 93\% of the original full-slide diagnostic performance \cite{Wang2026WSISum}. To support these operations, architectural innovations like 2DMamba replace quadratic attention with linear-complexity state space modeling, effectively "compressing" the computational complexity of gigapixel image analysis \cite{Zhang2024DMamba}.

Complementary to explicit token reduction, \textbf{Weakly Supervised Learning as Implicit Compression} identifies critical regions through attention mechanisms. CLAM utilizes clustering-constrained attention to localize diagnostically relevant subregions, effectively filtering out irrelevant background and stroma during slide-level modeling \cite{Lu2021DataEfficientCPath}. This is further specialized in PAMoE, which introduces pathology-aware mixture-of-experts routing to direct heterogeneous tissue patches to specialized experts while suppressing non-diagnostic content \cite{Wu2025MoEWSI}. Such dynamic multimodal gating mechanisms provide the theoretical foundation for fusing only the most essential evidence from complex, multi-source pathology data \cite{Cao2023MultiModal}.
To further improve sequence modeling, \textbf{Long Context} techniques \cite{Guo2025ContextMatters,li2024rethinking} have been applied to WSIs, enabling vision transformers to maintain global contextual understanding across extremely long sequences while reducing memory and computation overhead.

\subsection{Multimodal Instruction Learning and Cognitive-Level Compression}

The frontier of the field involves \textbf{Multimodal Instruction Learning and Cognitive-Level Compression}, where compression is achieved by distilling gigapixel images into concise, high-level clinical narratives. Quilt-LLaVA introduces localized narrative instruction data derived from histopathology videos, enabling models to perform cross-patch reasoning and summarize vast visual areas into key diagnostic insights \cite{Seyfioglu2024QuiltLLaVA}. Similarly, SlideChat provides a vision–language assistant capable of performing gigapixel-level pathology reasoning by training on large-scale WSI instruction datasets, effectively converting visual evidence into actionable dialogue \cite{Chen2025SlideChat}. To make these models deployable, CLOVER introduces a cost-effective instruction learning framework that freezes the backbone of large language models and trains only lightweight adapters, representing a form of parameter-level compression for complex pathology dialogue systems \cite{Chen2025InstructionPathology}.

Recent advances in efficient visual representation learning, including occlusion-based contrastive learning~\cite{yang2025one,feng2026efficient}, self-adaptive token bases~\cite{young2026fewer}, and robust spatial-concept alignment~\cite{young2026scalar}, have shown that visual features can be substantially compressed without sacrificing discriminative power. These principles have also proven effective in domain-specific tasks such as medical image interpretation~\cite{xu2023learning,yang2024segmentation,yang2023geometry}, pathology token compression~\cite{chen2026tc}, multimodal medical foundation models~\cite{xu2024medvilam,xu2024foundation} and vision-centric long-context compression~\cite{gao2026zerosense,he2026autoselect}, further motivating our capacity-constrained formulation for token pruning.

\begin{figure}[htbp]
    \centering
    \includegraphics[width=0.85\textwidth]{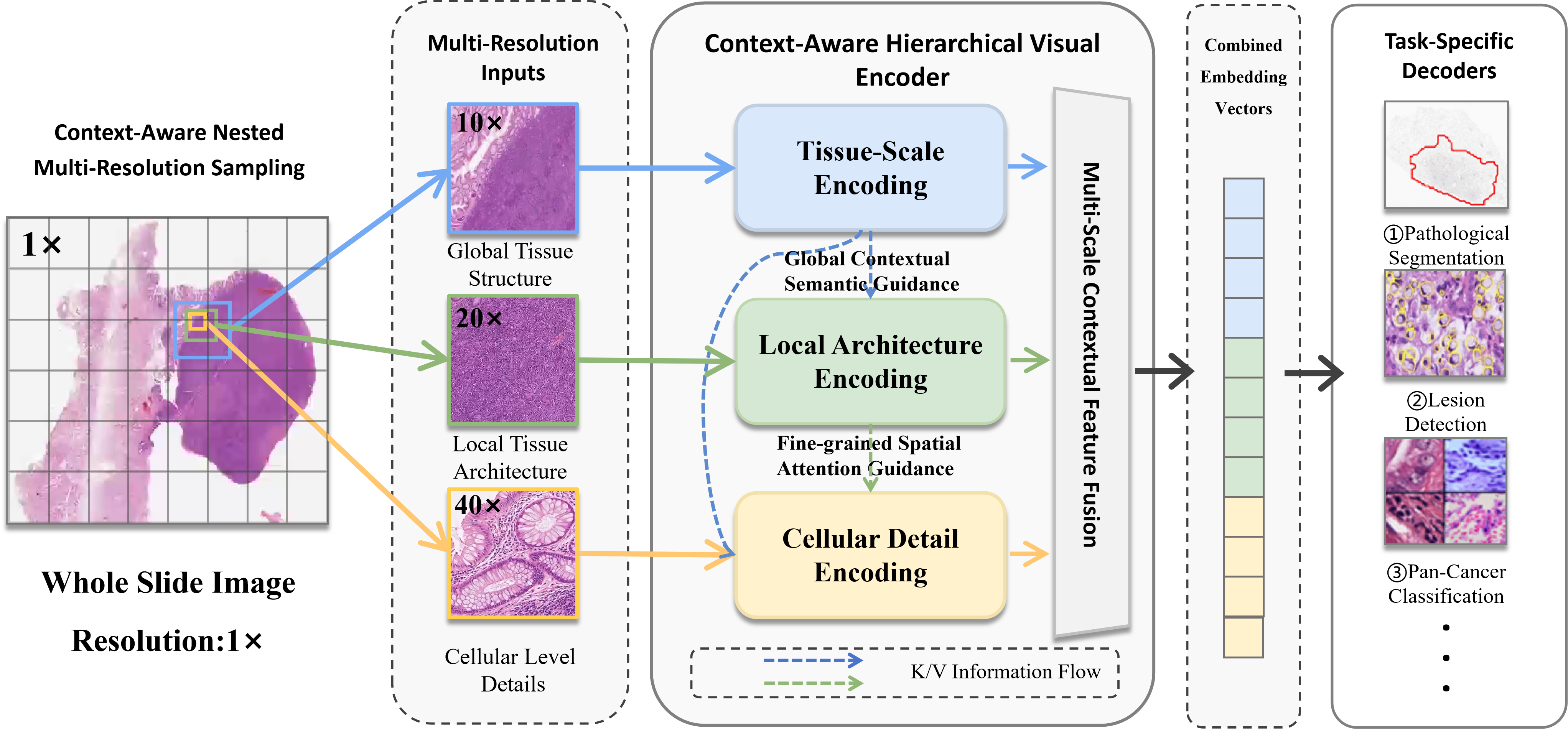}  % 请确保图片文件名正确
    \caption{Schematic diagram of adaptive multi-resolution context modeling for whole slide images. The framework constructs an image pyramid across magnifications (e.g., 5×, 10×, 20×, 40×) as multi-scale inputs. A hierarchical multi-resolution visual encoder captures global tissue context at low resolutions and fine cellular details at high resolutions. Cross-scale attention modules enable bidirectional information flow, where global tokens guide local detail extraction and local features refine global representations, building a complete context pyramid from cellular to tissue level.}
    \label{fig:multires_modeling}
\end{figure}

\section{Toward Clinically Actionable WSI Analysis: Scalable Perception, Data-Efficient Generalization, and Trustworthy Reasoning}

As pathology foundation models scale, research shifts from general-purpose pretraining to specialized clinical deployment. Real-world deployment faces three interconnected challenges: the computational burden of gigapixel WSIs, scarce annotations for rare diseases, and the need for transparent reasoning clinicians can trust. This section reviews recent advances addressing these challenges through multi-resolution context modeling, data-efficient adaptation, and interpretable multi-agent reasoning.

\subsection{Multi-Resolution Context Modeling for Gigapixel Images}

WSIs are inherently multi-scale, with diagnostic structures ranging from tissue architecture at low magnification to cellular details at high magnification. Early approaches processed WSIs by independently analyzing small patches, often losing spatial context. Recent work builds hierarchical representations that explicitly model cross-scale dependencies. A common strategy uses image pyramids: low-resolution branches capture global tissue organization, while high-resolution branches focus on cellular morphology. To fuse information across scales, architectures employ cross-scale attention mechanisms, where global tokens guide high-resolution detail extraction and local details refine global representations \cite{Guo2025ContextMatters, li2024rethinking}. This bidirectional flow builds coherent representations from nuclei to tissue architecture.

Beyond static fusion, adaptive computation has gained traction. Dynamic region selection mechanisms, inspired by pathologists' coarse-to-fine observation, allocate resources to diagnostically relevant areas. Importance scoring modules identify suspicious regions from low-resolution context and trigger fine-grained analysis only in those areas \cite{lin2025adaptvision, li2026token}, reducing computational burden while aligning model focus with clinical practice.

Recent work also explores compressing the reasoning process itself. LightThinker compresses intermediate reasoning into compact tokens during generation, reducing memory and latency by up to 70\% while preserving key information \cite{zhang2025lightthinker}. Parallel to this, adaptive multi-resolution modeling offers an efficient representation paradigm for WSI analysis. As shown in Figure~\ref{fig:multires_modeling}, a conceptual framework constructs a multi-scale input pyramid (e.g., 5× to 40× magnification). Low resolutions capture global architecture, high resolutions encode cellular details. A hierarchical encoder extracts features across scales, with cross-level attention enabling bidirectional interaction. An importance-aware module refines critical regions and compresses less relevant areas, ensuring efficiency without sacrificing diagnostic information.

\subsection{Multimodal Data Generation and Synthesis Efficiency}

Generative models provide an effective solution for addressing data scarcity in pathology. While initial efforts focused on diffusion models to synthesize realistic patches \cite{Zhang2022PseudoHealthy, Ding2023SyntheticPathologyDataset}, recent trends have shifted toward multimodal synthesis. 

\textbf{Multi-Agent Collaboration}, exemplified by PathGen \cite{sun2024pathgen}, leverages a dialogue between a ``Visual Agent'' and a ``Pathologist Agent'' to generate biologically plausible pathology image-text pairs, ensuring high-quality data for rare disease modeling. 
By simulating such a dialogue, these frameworks ensure that synthesized images remain biologically plausible and align with clinical terminology, providing high-quality data for training in rare disease contexts.

The efficiency of the generation process itself has also received increasing attention. FastFlow \cite{bajpai2026fastflow} addresses the inference bottleneck of Flow Matching models by framing the step-skipping decision as a multi-armed bandit problem, enabling adaptive computation during generation. This plug-and-play method achieves over 2.6× average acceleration without retraining, making large-scale synthesis of high-resolution pathology images more practical for augmenting training data in rare disease scenarios.

\begin{figure}[htbp]
    \centering
    \includegraphics[width=0.85\textwidth]{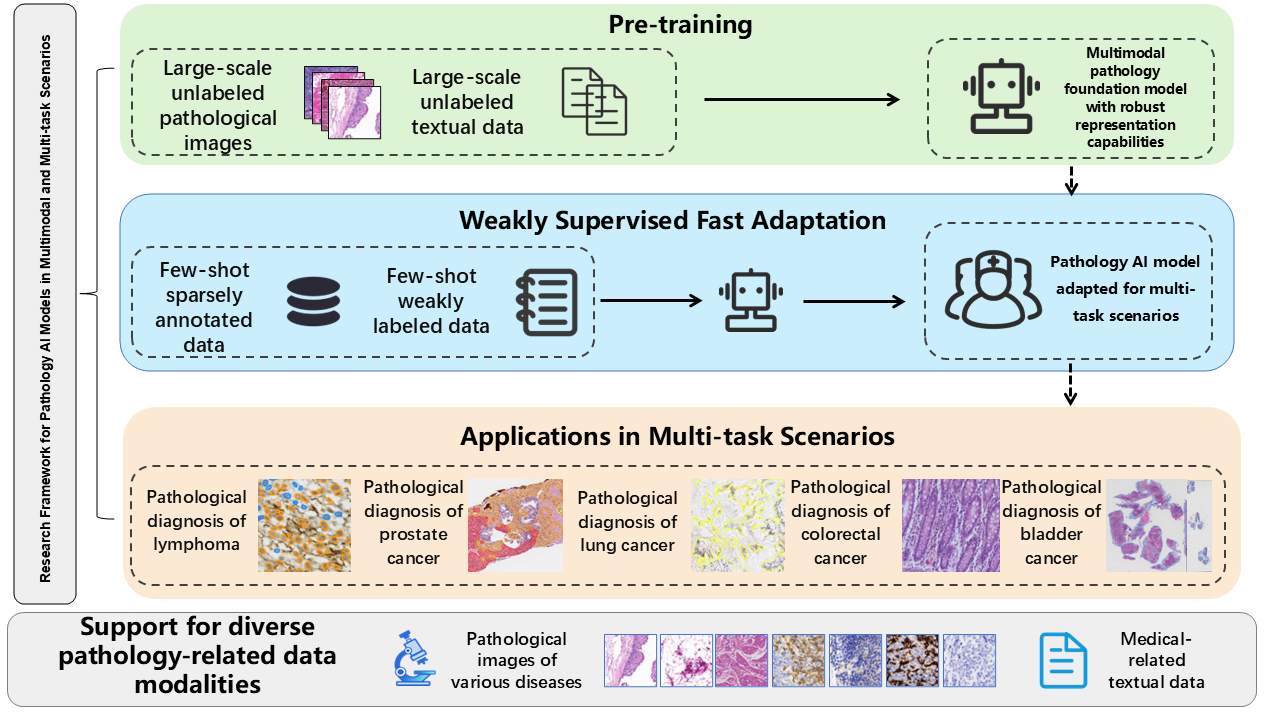}
    \caption{Schematic illustration of few-shot adaptation based on multimodal foundation models. Given a small support set of labeled pathology images and corresponding clinical descriptions, a pretrained multimodal model is adapted to new diagnostic tasks using parameter-efficient fine-tuning techniques (e.g., adapters, prompt tuning, or low-rank adaptation). The model leverages prior knowledge from large-scale pretraining to rapidly generalize to novel disease subtypes with only a few examples, significantly reducing the need for extensive expert annotations.}
    \label{fig:fewshot}
\end{figure}

\subsection{Parameter-Efficient Adaptation and Few-Shot Learning}
Few-shot learning is essential for adapting pathology models to ``long-tail'' clinical tasks where expert annotations are critically scarce. 
Figure~\ref{fig:fewshot} illustrates a conceptual framework for few-shot adaptation using multimodal foundation models.
Recent research focuses on parameter-efficient fine-tuning (PEFT) and prompting strategies to bridge the domain gap. Advanced foundation model adaptation techniques \cite{huang2024free, yin2024prompting} demonstrate that large-scale pathology models can be adapted to niche clinical tasks by updating only a small fraction of parameters, significantly improving efficiency and scalability. For instance, FOCUS introduces a knowledge-enhanced adaptive visual compression framework that leverages language priors for few-shot scenarios \cite{guo2025focus}.

Beyond parameter efficiency, recent work has explored how to fundamentally improve the few-shot learning capability of MLLMs for fine-grained recognition tasks. Fine-R1 \cite{he2026fine} addresses the challenge of high intra-class variance and low inter-class variance through a two-stage training framework: chain-of-thought supervised fine-tuning (CoT SFT) establishes structured reasoning patterns, followed by triplet augmented policy optimization (TAPO) that enhances robustness and discriminability using positive and negative image samples. With only 4-shot training, Fine-R1 achieves state-of-the-art performance on both seen and unseen fine-grained categories, outperforming even contrastive CLIP models. This demonstrates that improving knowledge deployment through structured reasoning and contrastive reinforcement can substantially boost few-shot recognition capability without requiring stronger visual features or external knowledge injection.

Beyond traditional classification, these adaptation strategies are being extended to complex clinical downstream tasks and biological integration. In the realm of prognosis, new frameworks for survival prediction integrate visual features with genomic data to provide interpretable scores, as demonstrated by the robust multimodal modeling of Zhou et al. \cite{zhou2025robust} and the interpretable architectures of Liu et al. \cite{liu2024interpretable}. Furthermore, AI is increasingly bridging the gap between morphology and molecular biology by predicting gene expression directly from WSIs \cite{nishimura2025learning, huang2025scalable}. This enables ``virtual'' molecular profiling, which saves critical time and resources in clinical workflows by extracting molecular insights from standard H\&E stained slides.

\begin{figure}[htbp]
    \centering
    \includegraphics[width=0.85\textwidth]{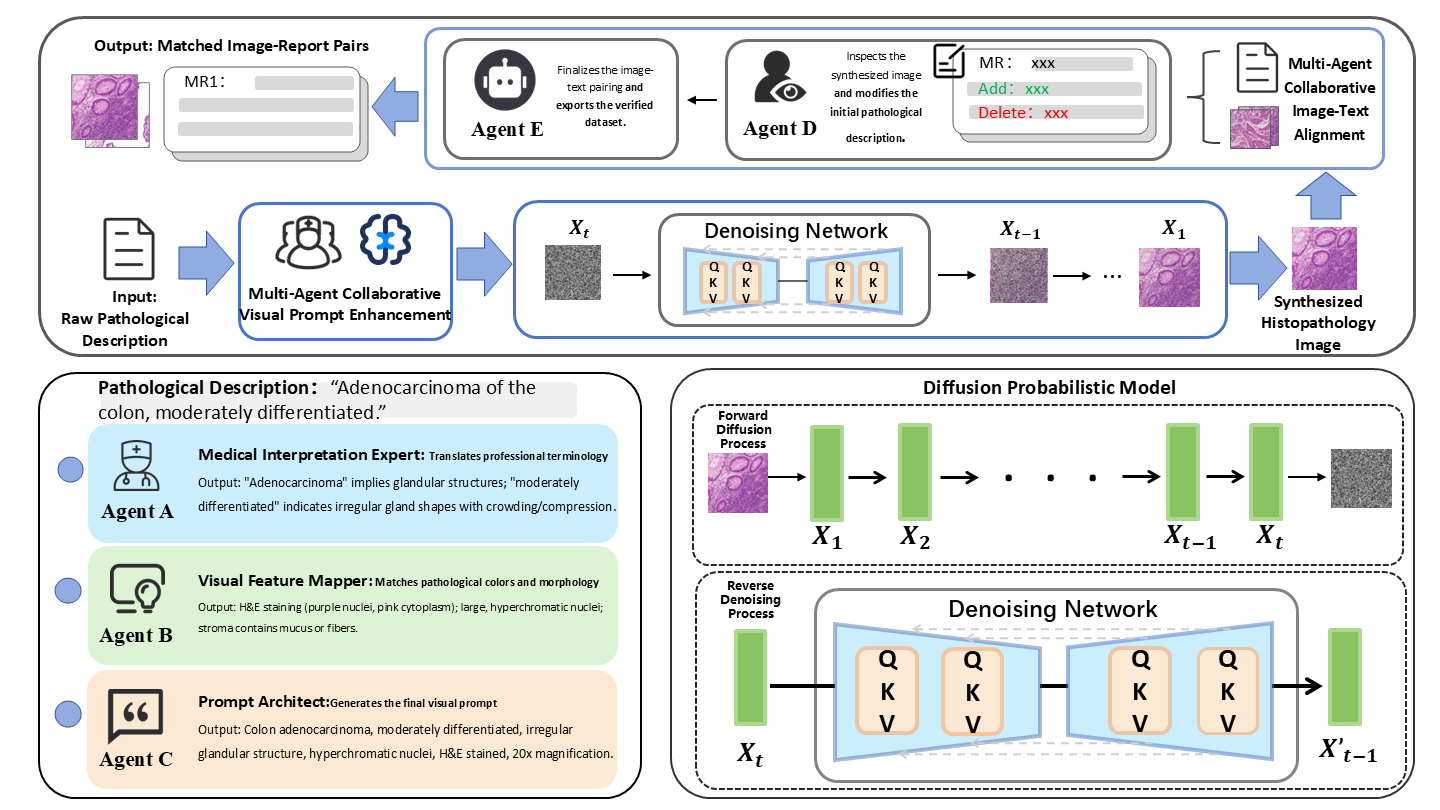}
    \caption{Schematic illustration of multi-agent collaborative framework for multimodal pathology image–text synthesis. Three specialized agents (medical interpreter, visual feature mapper, and prompt architect) collaboratively generate structured visual prompts from clinical descriptions. These prompts condition a diffusion model to synthesize high-fidelity pathology images, while a text generation module produces semantically consistent reports, ensuring cross-modal alignment. Synthesized samples undergo multi-discriminator assessment for image realism, textual accuracy, and image–text matching. Finally, collaborative training with curriculum learning and importance sampling integrates real and synthetic data to enhance model generalization, particularly for long-tail and rare disease scenarios.}
    \label{fig:generation_multi}
\end{figure}

\subsection{Multi-Agent Collaboration and Interpretable Reasoning}

Interpretability is essential for clinical AI, driving the development of reasoning frameworks beyond black-box models. Recent advances simulate pathologists' diagnostic processes via agent-based architectures. CPath-Agent employs multi-agent collaboration for systematic slide review, mirroring professional consultation logic \cite{Sun2025CPathAgent}. Figure~\ref{fig:generation_multi} illustrates how such collaboration emulates a pathologist's workflow. To address data imbalance and long-tail distributions, we propose a multi-agent framework for high-fidelity multimodal data synthesis. It establishes fine-grained image-text alignment through three specialized agents guiding conditional diffusion models, followed by multi-agent quality assessment and collaborative training to enhance generalization.

Beyond role design, training methodologies have advanced. Stronger-MAS introduces AT-GRPO, a multi-agent reinforcement learning algorithm enabling collaborative LLMs to dynamically improve teamwork via agent- and turn-wise reward grouping \cite{zhao2025stronger}, showing promise for training pathology agent teams that cross-validate findings.

Interpretability is further enhanced by architectural innovations and uncertainty quantification. MoE-WSI adapts mixture-of-experts for WSI analysis, using dynamic routing to assign tissue patterns to specialized experts, providing clear traceability of diagnostic features \cite{Wu2025MoEWSI}. Complementing this, Similarity-as-Evidence (SaE) reinterprets vision-language similarity as evidence and decomposes predictive uncertainty into vacuity and dissonance via Dirichlet modeling, enabling more reliable active learning and mitigating VLM overconfidence \cite{xie2026similarity}.

By integrating multi-agent collaboration, structured reasoning, and uncertainty-aware mechanisms, these frameworks move beyond black-box predictions toward trustworthy diagnostic assistance. Such systems articulate diagnostic logic, reflect on contradictory evidence, and alert clinicians to unreliable predictions, aligning with the pathologist's chain of thought and advancing toward clinically deployable AI assistants.

\section{Challenges and Future Research Directions}

Despite the transformative potential of multimodal AI in pathology, several critical hurdles must be overcome to achieve clinical maturity. The primary \textbf{Challenges} include: 1) \textbf{Efficient modeling of gigapixel WSIs}, where the quadratic complexity of standard attention mechanisms struggles with ultra-long visual sequences; 2) \textbf{Limited annotated datasets}, especially for rare disease subtypes and complex spatial phenotypes; 3) \textbf{Biological consistency in multimodal generation}, ensuring that synthetic images and reports maintain rigorous clinical accuracy; and 4) \textbf{Reliability and interpretability}, addressing the "black-box" nature of deep learning to ensure safe clinical decision-making. 

Looking forward, \textbf{Future Research Directions} are expected to gravitate toward: 1) \textbf{Unified pathology foundation models} that can seamlessly handle diverse tasks from segmentation to survival analysis within a single architecture; 2) \textbf{Efficient long-sequence architectures}, such as state-space models or linear-complexity transformers, to process entire slides without information loss; 3) \textbf{Multimodal biomedical knowledge integration}, which moves beyond simple image-text pairs to incorporate structured ontologies, spatial transcriptomics, and longitudinal electronic health records; and 4) \textbf{Human-AI collaborative diagnostic systems} that prioritize the "physician-in-the-loop" paradigm, fostering a synergistic relationship where AI enhances rather than replaces the pathologist's expertise.

\section{Conclusion}

This review has systematically examined recent progress in multimodal computational pathology, tracking the evolution from self-supervised representation learning and foundation models to large language models capable of reasoning across pathology images and clinical text. We highlighted the unique challenges posed by gigapixel whole-slide images, including extreme spatial resolution, sparse diagnostic information, and scale-dependent semantics. Key innovations such as structure-aware token compression, multi-resolution alignment, and reasoning-enhanced few-shot adaptation have enabled efficient processing and rapid deployment in low-data scenarios. Multi-agent collaborative reasoning frameworks have further advanced interpretability by mimicking the pathologist's diagnostic workflow and enabling uncertainty-aware evidence fusion.

Several challenges remain on the path to clinical adoption, including efficient modeling of ultra-long visual sequences, ensuring biological consistency in multimodal generation, and integrating structured biomedical knowledge such as ontologies and genomic profiles. Future progress will depend on unified pathology foundation models that span tasks from segmentation to prognostic reasoning, coupled with efficient long-sequence architectures and human-AI collaborative systems that prioritize the physician-in-the-loop paradigm. As these directions converge, multimodal computational pathology is poised to deliver scalable, trustworthy, and clinically impactful AI assistants that improve diagnostic accuracy and advance patient care.

% \clearpage
\bibliographystyle{unsrt}  
\bibliography{references} 

\end{document}